\documentclass{article}
\PassOptionsToPackage{numbers,compress}{natbib}

\usepackage[preprint]{tackling_climate_workshop_style}

\usepackage[utf8]{inputenc}
\usepackage[T1]{fontenc}
\usepackage{hyperref}
\usepackage{url}
\usepackage{booktabs}
\usepackage{siunitx}
\usepackage{amsmath}
\usepackage{amssymb}
\usepackage{graphicx}
\usepackage{microtype}
\usepackage{enumitem}
\usepackage{subcaption}
\usepackage{pdflscape}
\usepackage{algorithm}
\usepackage{algpseudocode}

\title{Online Reinforcement Learning in the Met Office Unified Model through Distributed Model–Agent Coupling}

\author{%
\parbox{\textwidth}{
\centering
Pritthijit Nath$^{1}$ \quad Sebastian Schemm$^{1}$ \quad Peter Haynes$^{1}$ \\ \vspace{0.1cm} Emily Shuckburgh$^{2}$ \quad Mark Webb$^{3}$} \vspace{0.2cm} \\
$^1$ Department of Applied Mathematics and Theoretical Physics, University of Cambridge\\
$^2$ Department of Computer Science and Technology, University of Cambridge\\
$^3$ Met Office Hadley Centre\\
\texttt{\{pn341,ss3299,phh1,efs20\}@cam.ac.uk}; \texttt{mark.webb@metoffice.gov.uk}
}

\begin{document}
\setcitestyle{square}
\maketitle

\vspace*{-5mm}

\begin{abstract}
Machine-learnt corrections can complement numerical weather prediction provided that they operate stably within an evolving numerical model. In this study, we couple the Met Office (UKMO) Unified Model (UM) with distributed reinforcement-learning agents through rank-local tensors. A column-aware deep deterministic policy gradient (DDPG) actor uses local vertical structure together with full-column context to apply bounded corrections to potential temperature and horizontal wind. During training, we perform ten nudged 6-hr 12-min forecasts, with nudging towards the UKMO operational analysis providing an immediate counterfactual target from which the policy learns. The resulting actor is then frozen and applied to a non-nudged forecast without access to analysis inputs or further weight updates. Relative to a matched non-nudged native forecast at +6~h, the corrected forecast reduces global latitude-weighted MAE by 2.85\% for $Z_{500}$, 2.16\% for MSLP, 5.16\% for $T_{500}$ and 2.27\% for $T_{1.5\textrm{m}}$, with an observed 3.57\% wall-time overhead compared to native execution. Even though training and inference share the same initialisation, this single-case experiment demonstrates significant promise and feasibility, laying the groundwork for RL-based bias correction and parametrisations within operational systems.

\end{abstract}

\enlargethispage{2\baselineskip}

\section{Introduction}

Artificial intelligence (AI) is increasingly used to accelerate weather prediction and represent processes that are difficult to resolve explicitly. Data-driven forecasting systems such as Aurora~\cite{bodnar_foundation_2025}, ESFM~\cite{ozdemir_esfm_2026}, FourCastNet~\cite{pathak_fourcastnet_2022}, Pangu-Weather~\cite{bi_panguweather_2023}, and GraphCast~\cite{lam_graphcast_2023} produce skilful forecasts at substantially lower inference cost than conventional numerical weather prediction (NWP) methods. Hybrid systems instead embed machine-learning (ML) components within NWP, retaining the resolved dynamics while learning selected unresolved tendencies. NeuralGCM~\cite{kochkov_neuralgcm_2024}, for example, demonstrates that learnt components can operate stably within evolving weather and climate model dynamics, with similar stability demonstrated in multi-week ICON aquaplanet integrations under appropriate physical constraints~\cite{bertoli_radiation_2025}. Such approaches could make kilometre-scale simulations and larger ensembles more affordable while improving the representation of unresolved processes. However, their credibility ultimately depends on remaining stable and physically consistent as external forcing pushes the climate beyond the training distribution.

Most ML-based corrections are learnt offline from fixed datasets and subsequently inserted into the numerical model without further adaptation. Strong offline performance therefore does not guarantee reliable coupled behaviour, and learnt corrections can destabilise the resolved state~\cite{bertoli_radiation_2025, rasp_deep_2018, brenowitz_spatially_2019}, although physical constraints and targeted treatments can improve coupled stability~\cite{bertoli_radiation_2025, yuval_stable_2021}. Online learning therefore offers a promising alternative, allowing corrections to adapt continually to the evolving model state and to regimes under-represented in the original training data while remaining embedded within the physical constraints of the numerical model.

\clearpage
\enlargethispage{2\baselineskip}

Reinforcement Learning (RL) provides one potential route to online learning because a policy (a learnt mapping from the current model state to an action) can act on the evolving model and optimise feedback from the resulting state. Idealised weather and climate experiments~\cite{nath_replacing_2026} show that RL can learn state-dependent corrections and parametrisation controls. Extending this approach to an established global model creates a design challenge where MPI/OpenMP Fortran code must synchronise with Python agents, exchange distributed three-dimensional fields, and transfer a learnt policy from analysis-informed training to non-nudged operational inference where future analyses are unavailable. We address this challenge by coupling the Met Office (UKMO) Unified Model (UM)~\cite{brown_unified_2012}, an operational forecasting system, to Python-based RL agents through SmartSim~\cite{partee_smartsim_2022}, an in-memory interface between simulations and ML services at high-performance-computing scale.

To the best of our knowledge, this work demonstrates the first execution of online RL within the distributed execution of an operational global NWP model. The key contributions are:
\begin{enumerate}[leftmargin=*,nosep]
    \item \textbf{An MPI rank-local online coupling} that exchanges UM state, actions, and rewards with distributed RL agents through SmartSim while respecting the model's domain decomposition.
    \item \textbf{A train-to-inference formulation} in which operational analysis nudging provides an immediate counterfactual target during training while the frozen policy is subsequently deployed without analysis nudging during inference.
    \item \textbf{A coupled feasibility evaluation} using learning progression, numerical stability, and standard meteorological verification metrics such as latitude-weighted mean absolute error (MAE) in near-surface temperature, mean sea-level pressure, and 500-hPa temperature and geopotential height.
\end{enumerate}

\section{Coupled-learning method}

\subsection{Distributed model--agent exchange}

We use an atmosphere-only UM N320 configuration with 40 km (approximate) horizontal resolution, 70 vertical model levels, and a 720-s timestep that also sets the model--agent exchange interval. The $640\times480$ horizontal grid is decomposed over a $12\times16 = 192$ tile layout, with each tile contained within a single MPI rank (shown in Figure~\ref{fig:app-rank-decomposition}). Each tile contains $30\times54 = 1,620$ atmospheric grid-cell columns. Matching Python ranks mirror this decomposition, with the temperature and wind-grid columns represented as synchronous vectorised environments. Fields remain on their native staggered grids with co-located pressure, avoiding global collection or horizontal interpolation for policy evaluation.

At each exchange, a UM rank writes its local state tensors to Redis through SmartRedis, the client library within SmartSim. The corresponding Python rank evaluates its shared policy for each field column and returns separate action tensors for potential temperature and the two horizontal-wind components. The UM applies these increments and publishes the reward diagnostics (defined in Section~\ref{sec:method-reward}). Cylc~\cite{oliver_workflow_2019} coordinates the entire workflow (Figure~\ref{fig:cylc-workflow}), including the Redis service, UM and agent tasks, ten training forecasts, checkpoint persistence and final inference. 12 independent Redis processes share one node, with rank $r$ assigned to process $p_i = r\bmod12$. The associated execution times are reported in \ref{app:profiling}.

\subsection{RL setup}
\label{sec:method-rl-setup}

\begin{figure}[t]
\centering
\begin{minipage}[c]{0.45\linewidth}
\centering
\includegraphics[width=\linewidth]{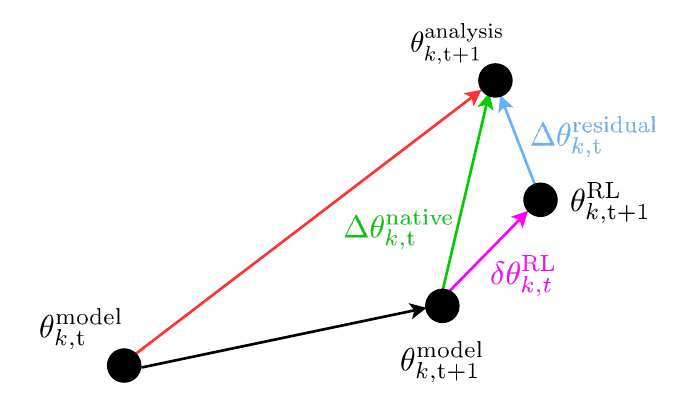}
\end{minipage}\hfill
\begin{minipage}[c]{0.50\linewidth}
\caption{Potential-temperature increments at level $k$ and exchange time $t$. Black nodes denote the model states $\theta^{\mathrm{model}}_{k,t}$ and $\theta^{\mathrm{model}}_{k,t+1}$, the RL-adjusted state $\theta^{\mathrm{RL}}_{k,t+1}$, and the analysis $\theta^{\mathrm{analysis}}_{k,t+1}$. The black and red arrows show the uncorrected model evolution and total model-to-analysis displacement, respectively. The green native increment $\Delta\theta^{\mathrm{native}}_{k,t}$ provides the counterfactual target. The magenta RL correction $\delta\theta^{\mathrm{RL}}_{k,t}$ changes the model state before nudging, leaving the blue residual increment $\Delta\theta^{\mathrm{residual}}_{k,t}$. Arrow lengths and directions are illustrative.}
\label{fig:increments}
\end{minipage}
\end{figure}

For each field $q\in\{\theta,u,v\}$, a column state contains seven 70-level profiles: normalised field value, one-step change, model level, vertical gradient, pressure, field identity and a temperature feature. The latter is model-derived air temperature for theta columns and zero for wind columns. The column-aware deep deterministic policy gradient (DDPG) actor~\cite{lillicrap_ddpg_2016} processes this 490-element state with vertical convolutions and a pooled full-column context, producing fractions $a_{q,k}\in[-1,1]$. Levels 1 and 70 are inactive, while levels 2--69 receive:
\begin{equation}
\delta q^{\mathrm{RL}}_k=\beta a_{q,k}\max(|q_k|,q_{\min}),
\qquad \beta=5.0\times10^{-5}.
\label{eq:action}
\end{equation}
Here $q_{\min}$ is 1~K for potential temperature and 1~m-s$^{-1}$ for wind, permitting corrections near zero wind. Figure~\ref{fig:increments} compares the nudging required before and after the policy action. Comparing the native increment $n_k=\Delta q_k^{\mathrm{native}}$ diagnosed before the action $\delta q^{\mathrm{RL}}_k$ with the residual increment $e_k=\Delta q_k^{\mathrm{residual}}$ required afterwards measures how much the policy reduces the remaining correction. Temperature diagnostics use air-temperature units through Exner conversion, and wind diagnostics retain wind-speed units.

\clearpage

\enlargethispage{2\baselineskip}

Each MPI-matched rank holds its own actor, critic, target networks and replay buffer, with weights shared across fields and columns on that rank but not across MPI ranks. Both networks use 32-channel vertical convolutions and pointwise heads. The replay buffer stores complete columns, retains at most 32 columns per field per exchange and samples 32-column batches balanced across the three fields and five pressure bands separated at 200, 400, 600 and 850~hPa. Each eligible exchange performs 16 critic updates and eight actor updates on CPUs. A training-only squared target-action term with coefficient 0.02 supplements the actor objective. Table~\ref{tbl:implementation} and Algorithm~\ref{alg:app-pseudocode-DDPG} give the configuration and update sequence.

\subsection{Analysis-informed reward and non-nudged inference}
\label{sec:method-reward}

For active levels $A=\{2,\ldots,69\}$, define layer mass $m_k=\Delta p_k/g$ from layer-pressure thickness $\Delta p_k$ and gravitational acceleration $g$. The mean-normalised mass is $\widetilde m_k=m_k/\overline m_A$, where $\overline m_A=|A|^{-1}\sum_{k\in A}m_k$. With $\langle\cdot\rangle_A$ denoting the active-level mean, the level and column scores are:
\begin{equation}
s_k=\frac{n_k^2-e_k^2}{n_k^2+e_k^2+\epsilon_{32}},\qquad
C=\frac{\langle\widetilde m n^2\rangle_A-\langle\widetilde m e^2\rangle_A}
{\langle\widetilde m n^2\rangle_A+\langle\widetilde m e^2\rangle_A+\epsilon_{32}},
\qquad r_k=0.9s_k+0.1C.
\label{eq:reward}
\end{equation}
The single-precision machine epsilon $\epsilon_{32}$ prevents division by zero in the numerical diagnostic units. Positive $s_k$ denotes a smaller squared residual correction. Without RL corrections, $n_k=e_k$ and both scores are zero. The column term compares mass-weighted squared corrections, preventing cancellation of opposite-signed errors. Both scores ($C$ and $s_k$) and their convex combination $r_k$ are bounded by one in magnitude. The critic learns from $r_k$, while reported column rewards are $r=\langle r_k\rangle_A$. Fields are scored separately, and the reward curve (Figure~\ref{fig:temp-reward-progression}) reports temperature columns.

During training, the UM applies the RL correction $\delta q^{\mathrm{RL}}_k$ before nudging, as shown in Figure~\ref{fig:cylc-workflow}. Under inference, nudging is disabled, although the same counterfactual target is diagnosed for evaluation. The analysis is neither supplied to the actor nor applied to the prognostic state. Exploration, replay insertion, and weight updates are likewise disabled, leaving only the frozen actor to modify the forecast.

\section{Results}

\subsection{Transfer to non-nudged execution}

\begin{figure}[t]
\vspace{-6mm}
\centering
\begin{minipage}[c]{0.45\linewidth}
\centering
\includegraphics[width=\linewidth]{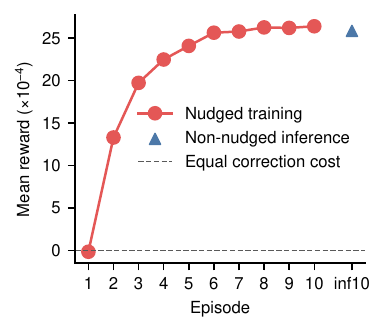}
\end{minipage}\hfill
\begin{minipage}[c]{0.53\linewidth}
\caption{Temperature-reward progression. The ordinate is the rank-time mean reward in units of $10^{-4}$. Connected red circles denote ten nudged training episodes, while the separate blue triangle at {inf10} denotes non-nudged inference from the episode-ten checkpoint. The dashed zero line marks equal native and residual correction costs, not a measured reward from the native control. Rank-time summaries are dependent diagnostics, not independent forecasts.}
\label{fig:temp-reward-progression}
\end{minipage}
\vspace{-2mm}
\end{figure}

The workflow completed ten nudged training forecasts and one frozen non-nudged inference, each initialised at 00 UTC on 12 December 2021 and lasting 6-hr 12-min. Each forecast contained 31 state--action exchanges at 12-min intervals and 30 rewarded transitions, giving $192\times30=5{,}760$ temperature rank-time reward summaries. Their equally weighted mean (Figure~\ref{fig:temp-reward-progression}) rose from $-1.72\times10^{-5}$ in the first forecast to $2.64\times10^{-3}$ in the tenth, with $2.58\times10^{-3}$ during inference. Learning was not strictly monotonic, with a small decline at episode nine. Restoring the checkpoint with nudging, exploration and learning disabled completed the transfer to non-nudged execution. With Dr Hook~\cite{saarinen_drhook_2005} (UM profiler) disabled in both forecasts, the inference UM task takes 116-s against 112-s for native execution, an observed overhead of 3.57\% (\ref{app:profiling}).

\clearpage
\enlargethispage{2\baselineskip}

\begin{table}
\centering
\caption{Forecast-error changes relative to the non-nudged native control. At +6~h, $\Delta$MAE is coupled RL minus native MAE against UKMO analysis, and percentage change is $100\Delta\mathrm{MAE}/\mathrm{MAE}_{\mathrm{native}}$. Negative values indicate improvement. Analysis time aggregation is specified in Section~\ref{sec:reference-diagnostics}. Means use cosine-latitude weights, and percentages are calculated before display rounding.} \vspace{2mm}
\label{tbl:zonal-verification}
\footnotesize
\resizebox{\linewidth}{!}{%
\begin{tabular}{@{}rrrrrrrrr@{}}
\toprule
& \multicolumn{4}{c}{Near-surface diagnostics} & \multicolumn{4}{c}{500-hPa diagnostics}\\
\cmidrule(lr){2-5}\cmidrule(lr){6-9}
Latitude
& \multicolumn{2}{c}{$T_{1.5\mathrm{m}}$ ($^\circ$C)}
& \multicolumn{2}{c}{MSLP (hPa)}
& \multicolumn{2}{c}{$T_{500}$ ($^\circ$C)}
& \multicolumn{2}{c}{$Z_{500}$ (m)}\\
\cmidrule(lr){2-3}\cmidrule(lr){4-5}\cmidrule(lr){6-7}\cmidrule(lr){8-9}
& $\Delta$MAE & Change (\%)
& $\Delta$MAE & Change (\%)
& $\Delta$MAE & Change (\%)
& $\Delta$MAE & Change (\%)\\
\midrule
Global & $-0.012$ & $-2.27\%$ & $-0.009$ & $-2.16\%$ & $-0.012$ & $-5.16\%$ & $-0.077$ & $-2.85\%$\\
\midrule
60--90$^\circ$N & $-0.022$ & $-1.37\%$ & $-0.021$ & $-4.52\%$ & $-0.021$ & $-9.02\%$ & $-0.127$ & $-4.93\%$\\
30--60$^\circ$N & $-0.015$ & $-1.97\%$ & $-0.004$ & $-0.69\%$ & $-0.010$ & $-4.06\%$ & $-0.125$ & $-4.35\%$\\
0--30$^\circ$N & $-0.008$ & $-1.68\%$ & +0.009 & +2.63\% & $-0.011$ & $-5.01\%$ & $-0.015$ & $-0.69\%$\\
0--30$^\circ$S & $-0.011$ & $-2.87\%$ & $-0.027$ & $-6.53\%$ & $-0.011$ & $-4.38\%$ & $-0.155$ & $-6.44\%$\\
30--60$^\circ$S & $-0.016$ & $-5.92\%$ & $-0.014$ & $-4.06\%$ & $-0.007$ & $-2.96\%$ & $-0.002$ & $-0.07\%$\\
60--90$^\circ$S & $-0.006$ & $-1.24\%$ & +0.000 & +0.05\% & $-0.034$ & $-13.15\%$ & $-0.041$ & $-0.83\%$\\
\bottomrule
\end{tabular}%
}
\vspace{-3mm}
\end{table}

\subsection{Forecast comparison across variables and latitude}

As the reward need not reflect forecast quality across all variables, we compare +6-h forecasts with UKMO analysis. 500-hPa geopotential height ($Z_{500}$) and mean sea-level pressure (MSLP) references are hourly means, while 1.5-m temperature ($T_{1.5\mathrm{m}}$) uses hourly maximum (Section~\ref{sec:reference-diagnostics}). 500-hPa temperature ($T_{500}$) uses the instantaneous value. Relative to the non-nudged native control from the same initial state, Table~\ref{tbl:zonal-verification} shows global latitude-weighted MAE reductions of 2.85\%, 2.16\%, 5.16\% and 2.27\% for $Z_{500}$, MSLP, $T_{500}$ and $T_{1.5\mathrm{m}}$ respectively. $T_{500}$ and $Z_{500}$ improve in every band, whereas MSLP deteriorates in the northern tropics and southern polar band. Expanded MAEs, spatial errors and biases appear in \ref{app:spatial}. Comparisons against the analysis-nudged control are presented in \ref{app:nudged-reference}.

\section{Discussion and conclusion}

Building on Nath et al.~\cite{nath_replacing_2026}, this study demonstrates distributed online learning and frozen non-nudged inference within a global NWP model. Fortran UM ranks and RL agents exchange decomposed tensors and preserve learning state across forecasts. The diagnosed increment retains its counterfactual meaning in both phases, with inference using analysis only for diagnostics and verification. The four headline global MAEs improve relative to non-nudged native execution, although regional MSLP and signed-bias changes are not uniformly beneficial. Temperature errors remain larger than in the analysis-nudged control, which receives future analysis information during integration.

The rewards use relative squared correction errors, whereas verification uses absolute errors after coupled forecast evolution. A positive immediate reward therefore does not guarantee improvement in every diagnostic. Training and configuration selection used the same initialisation, hence the measured gains cannot establish generalisation. Future experiments should separate selection from held-out dates and retain matched non-nudged and analysis-nudged controls, together with energy and mass budgets. Within the current evidence boundary, the successful transfer of a state-dependent online policy to non-nudged inference in a distributed NWP setting demonstrates significant promise and provides a practical basis for evaluating RL-derived bias correction and parametrisations of unresolved sub-grid processes.

\clearpage

\begin{ack}
P. Nath was supported by the \href{https://ai4er-cdt.esc.cam.ac.uk/}{UKRI Centre for Doctoral Training in Application of Artificial Intelligence to the study of Environmental Risks} [EP/S022961/1]. Mark Webb was supported by the Met Office Hadley Centre Climate Programme funded by DSIT. The authors thank the Met Office for providing access to the Unified Model, analysis data, and the Met Office EX computing platform under a CASE studentship agreement with the University of Cambridge. We thank Andrew Shao and Alessandro Rigazzi (HPE) for their valuable assistance with setting up SmartSim.
\end{ack}

\section*{LLM usage disclosure}
The authors acknowledge the use of AI language models, specifically OpenAI Codex and ChatGPT~(GPT-5.6-sol and GPT-6-astra), during the preparation of this work. These tools were used to assist in software development, polish language usage and improve the overall clarity of the manuscript. All AI-generated content was reviewed, verified, and edited by the authors to ensure accuracy and appropriateness.

\small
\bibliographystyle{vancouver}
\bibliography{bibliography}

\normalsize

\appendix
\renewcommand\thesection{Appendix \Alph{section}}
\renewcommand\thesubsection{\Alph{section}.\arabic{subsection}}
\renewcommand\thetable{\Alph{section}.\arabic{table}}
\renewcommand\thefigure{\Alph{section}.\arabic{figure}}
\renewcommand\theHfigure{\Alph{section}.\arabic{figure}}
\setcounter{table}{0}
\setcounter{figure}{0}

\section{Implementation details}
\label{app:implementation}
\setcounter{figure}{0}

Algorithm~\ref{alg:app-pseudocode-DDPG} summarises the column-aware DDPG updates. Table~\ref{tbl:implementation} collects the state transformations, action range and learning settings. Potential temperature is mapped logarithmically, while wind components and one-step changes use field-specific linear scales. Pressure is bounded by $p_{\mathrm{ref}}=1{,}100$~hPa and defines the upward coordinate $-\log(p/p_{\mathrm{ref}})$ used for the vertical gradient of the normalised field. A 0.01-hPa floor keeps the logarithm finite at the model top. The temperature feature is $T=\theta\Pi$, where $\Pi=(p/1{,}000\,\mathrm{hPa})^{287.05/1005}$, and is zero for wind columns rather than interpolating temperature onto their grids. All features are derived from the model, not the verification analysis.

\begin{table}[!h]
\centering
\small
\caption{State transformations, action range and column-aware DDPG settings. Each field column supplies seven 70-level profiles.}
\vspace{2mm}
\label{tbl:implementation}
\begin{tabular}{lll}
\toprule
Quantity & Physical or numerical range & Agent representation\\
\midrule
Potential temperature $\theta_k$ & 150--6000 K & logarithmically mapped to $[-1,1]$\\
One-step tendency $\Delta\theta_k$ & $-10$--10 K & linearly mapped to $[-1,1]$\\
Wind component $u_k$ or $v_k$ & $-150$--150 m s$^{-1}$ & linearly mapped to $[-1,1]$\\
One-step wind change & $-20$--20 m s$^{-1}$ & linearly mapped to $[-1,1]$\\
Model level $k$ & 1--70 & linearly mapped to $[-1,1]$\\
Pressure $p_k$ & 0--1,100 hPa & linearly mapped to $[-1,1]$\\
Vertical gradient & clipped to $[-1,1]$ & gradient of normalised field\\
Field identity & $\theta$, $u$, $v$ & $-1$, 0, 1\\
Air-temperature feature & 100--400 K & mapped to $[-1,1]$, zero for winds\\
Action $a_k$ & $[-1,1]$ & levels 2--69 active\\
Actor and critic & 32-channel convolutions & local and full-column context\\
Replay & 2,450 columns per field & 32 retained per field per exchange\\
Optimisation batch & 32 complete columns & balanced fields and pressure losses\\
Updates per eligible exchange & 16 critics, 8 actors & discount 0.99\\
Target-action supervision & coefficient 0.02 & training only, all three fields\\
\bottomrule
\end{tabular}
\end{table}

\begin{landscape}
\begin{figure}[p]
\centering
\includegraphics[width=0.94\linewidth]{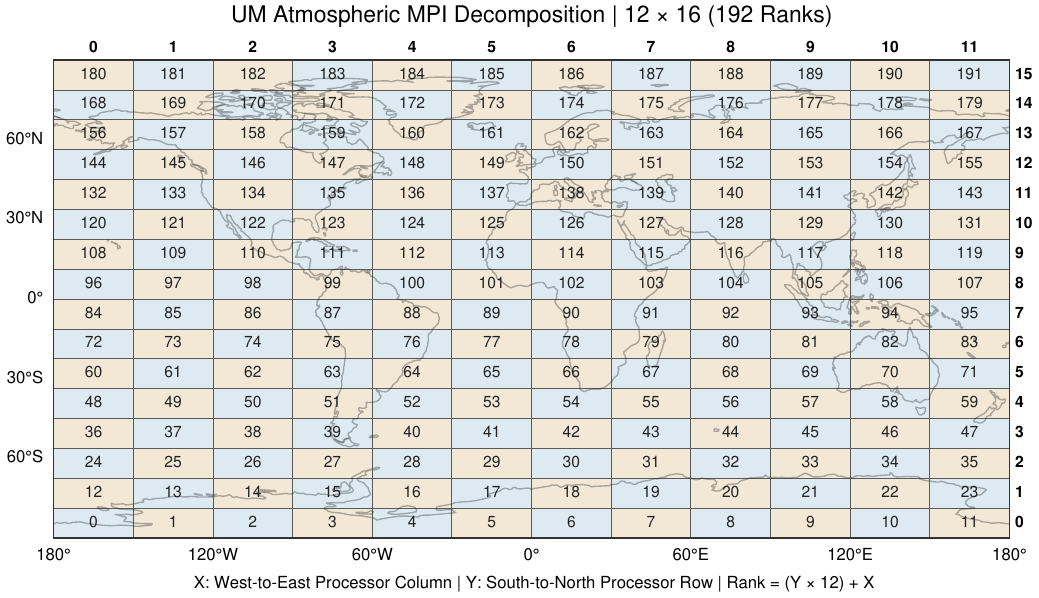}
\caption{Horizontal MPI decomposition of the $640\times480$ N320 atmosphere grid. The $12\times16$ processor layout contains 192 ranks, numbered from south to north and west to east. Each rank owns 30 rows and either 53 or 54 columns, giving 1,590 or 1,620 atmospheric columns. A single Python agent rank manages the vectorised environments for each tile.}
\label{fig:app-rank-decomposition}
\end{figure}
\end{landscape}

The coupling follows the decomposition in Figure~\ref{fig:app-rank-decomposition}. Each network has two 32-channel vertical convolutions with kernel width three and dilations one and two. Their local features are concatenated with the vertically averaged encoded column, followed by two 32-channel pointwise layers and a scalar head at each level. The critic additionally receives the column action. Each rank retains 7,350 complete field columns, corresponding to 499,800 active-level transitions, from the nominal 500,000-level capacity. Inference evaluates at most 128 columns per actor batch and restores only actor weights, without allocating replay, critics or optimisers.

For a batch, loss weights $\omega_{i,k}$ sum to one and give equal total weight to each represented field and each occupied pressure band within that field. The actor minimises $-\sum\omega_{i,k}Q_{i,k}+0.02\sum\omega_{i,k}(a_{i,k}-a^*_{i,k})^2$, where $a^*$ is the diagnosed native correction divided by the pre-action unit-action scale and clipped to $[-1,1]$. The theta scale includes Exner conversion to temperature-diagnostic units. This target is used only during training. Updates begin after ten exchanges, and replay retains complete columns so that vertical context is preserved rather than treating levels as independent network inputs.

\begin{algorithm}[!h]
\caption{Column-aware DDPG with field-balanced replay}
\label{alg:app-pseudocode-DDPG}
\begin{algorithmic}[1]
\State \textbf{Input:} active levels $A$, fields $F$, pressure bands $P$, discount $\gamma$, target coefficient $\tau$, batch size $B$, update count $G$, and exploration noise $\epsilon$
\State \textbf{Initialise:} column actor $\pi_\psi$, level-output column critic $Q_\phi$, target networks $\pi_{\psi'}$ and $Q_{\phi'}$, and field-balanced replay $\{\mathcal D_q\}_{q\in F}$
\For{each model--agent exchange $t$}
    \State Receive states $x_c\in\mathbb{R}^{7\times70}$ for all local field columns $c$
    \State Select bounded column actions $a_c=\pi_\psi(x_c)+\epsilon_c$ and zero inactive levels
    \State Observe $x'_c$, level rewards $r_{c,k}$, targets $a^*_c$ and physical termination $d_c$
    \State Retain up to 32 complete transitions per field in $\mathcal D_q$
    \If{$t>\text{learning\_starts}$}
        \For{$g=1$ \textbf{to} $G$}
            \State Sample $B$ columns approximately equally across fields and form weights $\omega_{i,k}$
            \State $y_{i,k}\gets r_{i,k}+\gamma(1-d_i)Q_{\phi',k}(x'_i,\pi_{\psi'}(x'_i))$
            \State Update $\phi$ by minimising $\sum_{i,k}\omega_{i,k}(Q_{\phi,k}(x_i,a_i)-y_{i,k})^2$
            \State Every second critic update, update $\psi$ using the actor loss defined above
            \State At each actor update, soft-update both target networks with coefficient $\tau$
        \EndFor
    \EndIf
\EndFor
\end{algorithmic}
\end{algorithm}

\clearpage

\section{Coupling workflow}
\label{app:coupling-workflow}
\setcounter{figure}{0}

\begin{figure}[!h]
\centering
\makebox[\linewidth][c]{\includegraphics[width=1.45\linewidth]{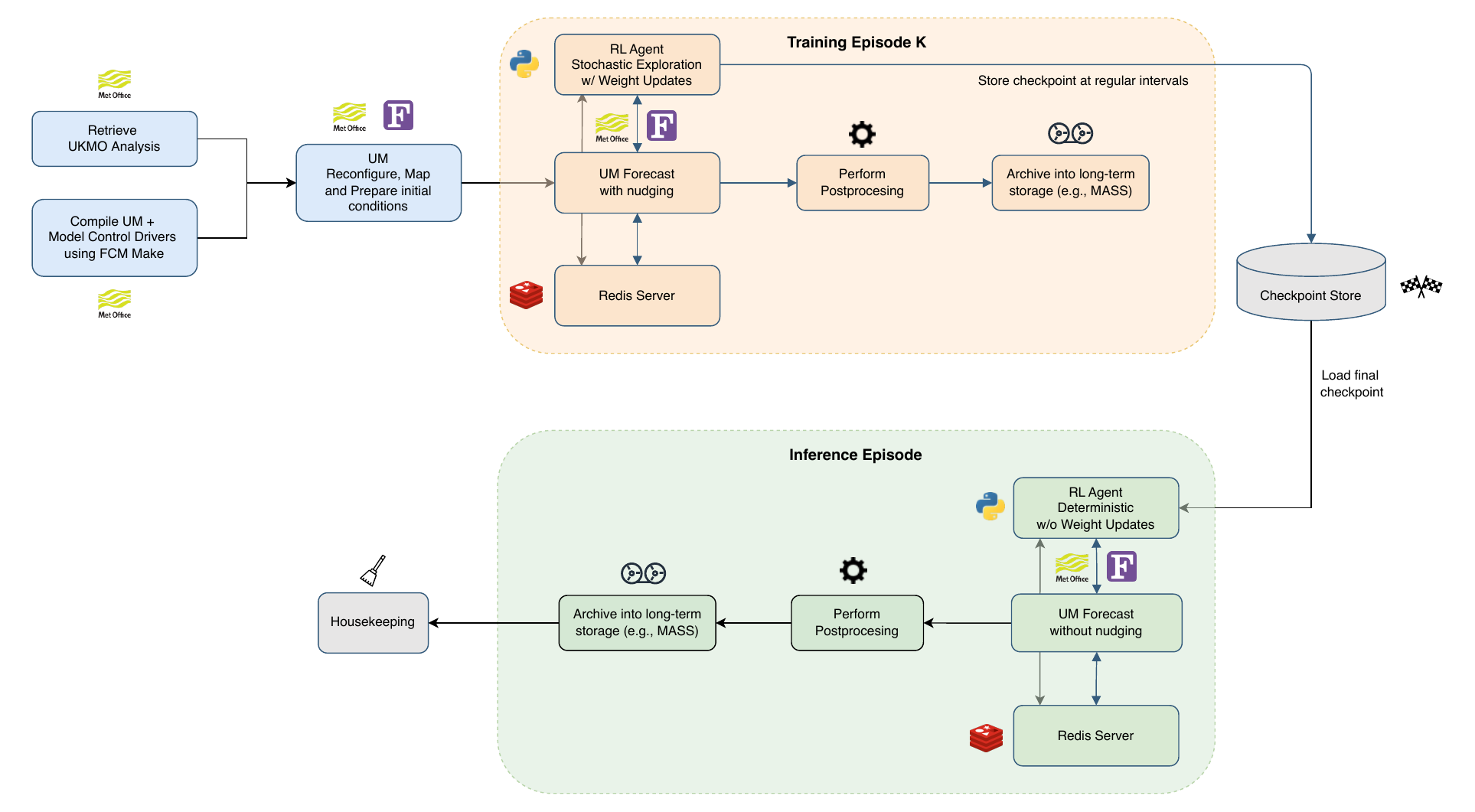}}
\caption{Cylc workflow for training and inference. Each nudged training forecast runs the UM, Redis service and RL-agent tasks together and writes learning state to the checkpoint store. In this experiment, saving occurs at the end of each training forecast, with only the latest checkpoint retained. Inference restores the final actor, disables exploration and updates, and runs without analysis nudging. Post-processing and archival remain outside the model--agent exchange.}
\label{fig:cylc-workflow}
\end{figure}

Figure~\ref{fig:cylc-workflow} shows how the workflow separates forecast orchestration, timestep communication, and learning-state persistence. Cylc controls forecast-level task dependencies, Redis handles timestep-level tensor exchange between the UM and agents, and checkpoints carry the learnt state between successive forecasts.

\begin{figure}[!h]
\centering
\makebox[\linewidth][c]{\includegraphics[width=1.15\linewidth]{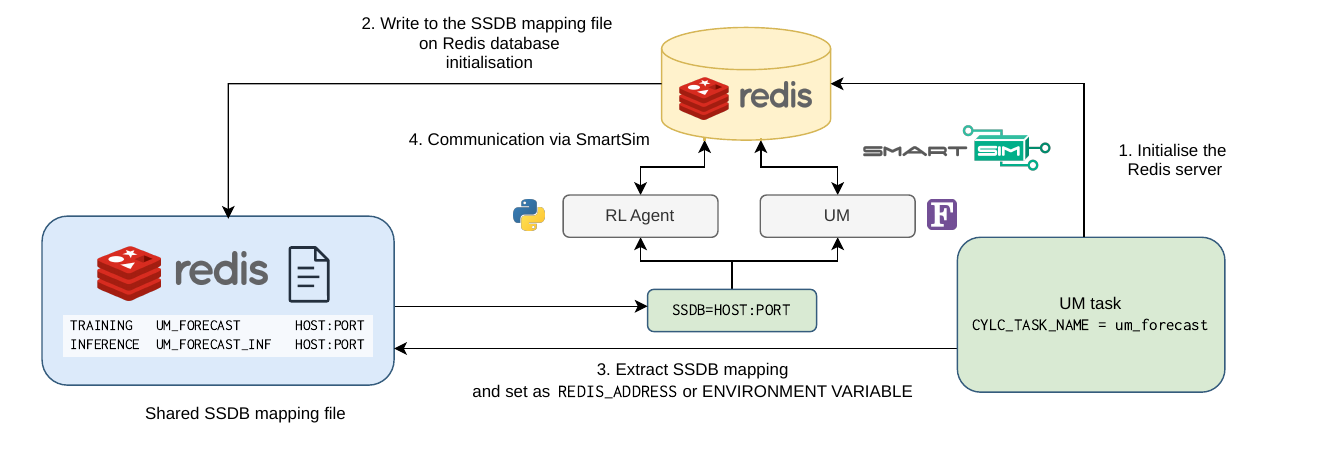}}
\caption{Redis discovery through the shared SSDB mapping, shown schematically for one endpoint. The measured configuration publishes 12 endpoints on one node, and matching UM and agent ranks select the same endpoint by rank modulo 12. Separate task names distinguish training from inference. This mapping avoids a fixed address because the scheduler can place the service on a different host for each forecast.}
\label{fig:app-ssdb-wiring}
\end{figure}

The PBS scheduler can place the Redis service on a different host for each forecast. To avoid embedding a database address in either coupled component, the workflow records the assigned host and ordered list of 12 ports in the shared SSDB mapping. The UM and agent resolve the same rank-assigned address using the current task name. Figure~\ref{fig:app-ssdb-wiring} shows this logical discovery mechanism, with the Redis service implemented as independent processes on one node. The task name changes between training and inference, but rank-local ownership and the tensor protocol do not.

\begin{figure}[!h]
\centering
\makebox[\linewidth][c]{\includegraphics[width=1.375\linewidth]{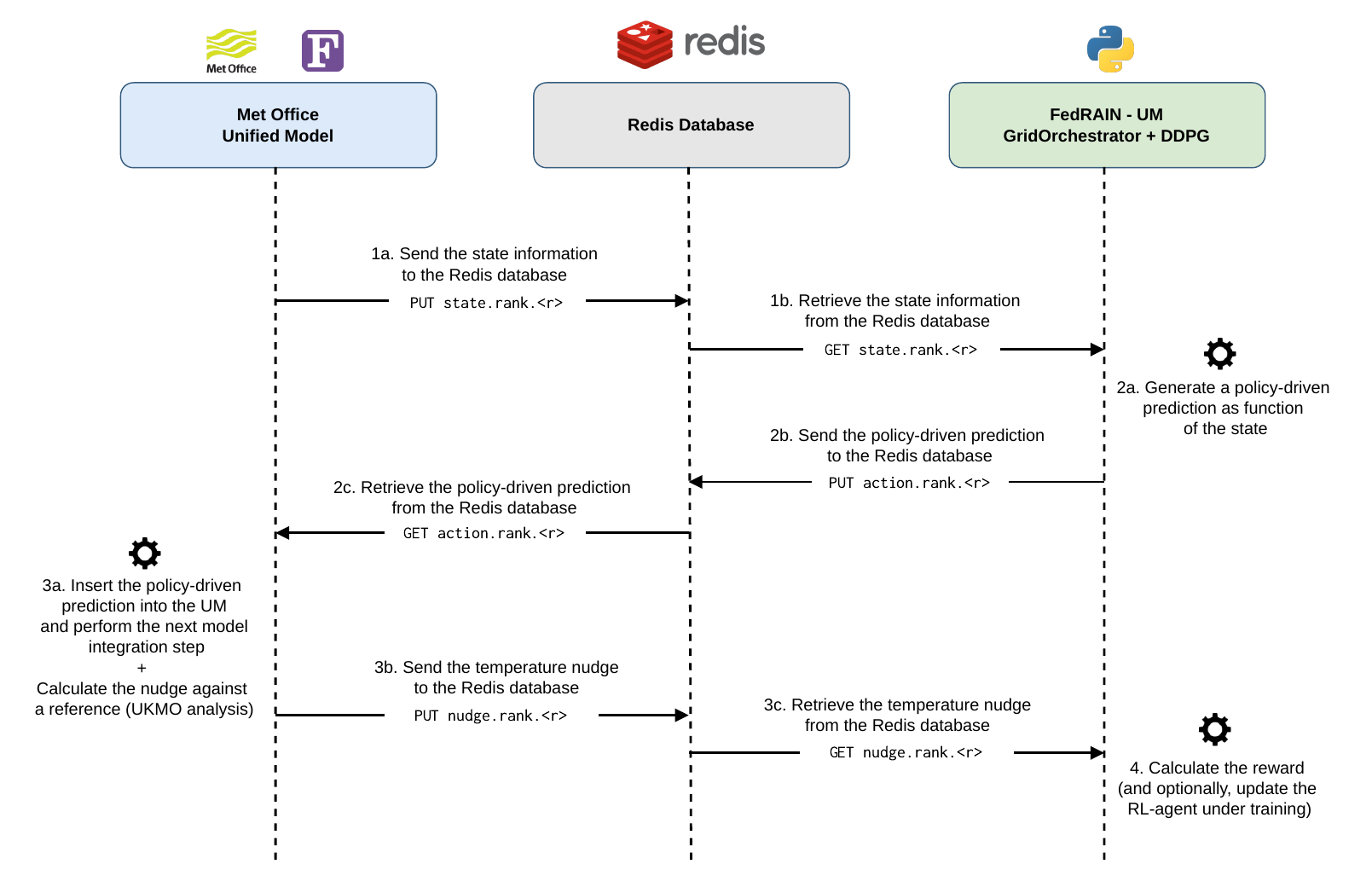}}
\caption{Rank-local UM--agent exchange for MPI rank $\langle r\rangle$, illustrated for the temperature path. The UM publishes local state, the matching agent returns an action, and the subsequent diagnostic enables reward calculation and optional training. The measured configuration repeats this protocol for both wind components and exchanges native and residual correction diagnostics. Rank-qualified keys preserve ownership and the association between states, actions and diagnostics.}
\label{fig:app-tensor-exchange}
\end{figure}

Communication is synchronised at two timescales. Within each forecast, exchanges follow a state--action--diagnostic sequence: every UM rank waits for its matching agent to return an action, then publishes the native and residual increments used to evaluate it. Rank-qualified keys and delete-after-consumption polling prevent reuse of stale actions or diagnostics. The existing UM MPI exchanges retain responsibility for communication between neighbouring model subdomains. Figure~\ref{fig:app-tensor-exchange} illustrates the rank-local protocol, which is repeated for temperature and both wind components.

\begin{figure}[!h]
\centering
\includegraphics[width=0.85\linewidth]{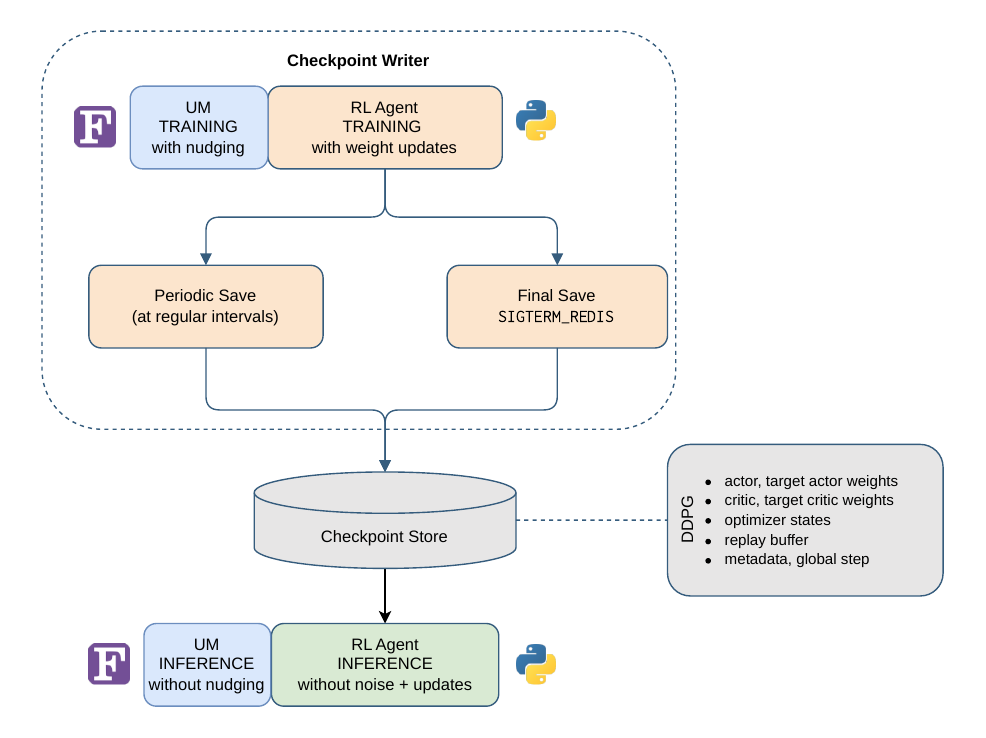}
\caption{Checkpoint transfer between training and inference. The framework supports periodic and final saves, while the measured configuration saves once per completed training forecast and retains only the latest checkpoint. Training restores actor, critic, target networks, optimisers, replay and metadata. Inference loads only the final actor weights, without constructing learning-only objects or updating parameters.}
\label{fig:app-checkpoint-flow}
\end{figure}

Across forecasts, the workflow persists the actor, critic, target networks, optimiser, replay buffer, metadata, and global step, as shown in Figure~\ref{fig:app-checkpoint-flow}. Subsequent training forecasts therefore resume the accumulated optimisation trajectory rather than restarting, whereas inference restores the final actor with exploration, replay insertion, and gradient updates disabled, making each action a deterministic function of the saved policy and current UM state.

\enlargethispage{2\baselineskip}

\clearpage

\clearpage

\section{Coupling profiling metrics}
\label{app:profiling}
\setcounter{table}{0}
\setcounter{figure}{0}

We measure complete task durations in a ten-forecast training sequence followed by frozen inference, with Dr Hook disabled in both the coupled experiment and the native control. Python timers remain enabled at the model--agent boundary, while a separate diagnostic experiment also enables Dr Hook within the UM. Table~\ref{tbl:app-profiling} reports diagnostic-experiment timings as the mean and population standard deviation of rank-local mean invocation times across all 192 agent ranks. In both experiments, the final training and inference agent tasks used four Genoa nodes, with 48 MPI ranks per node and two CPU threads per rank. The corresponding UM tasks used 192 MPI ranks with two OpenMP threads per rank on two Genoa nodes. 12 independent Redis processes shared a separate Genoa node. Task durations include initialisation, partner synchronisation and shutdown but exclude scheduler queueing, compilation, post-processing and archival.

\begin{table}[!h]
\centering
\small
\caption{Complete PBS task wall times for 6-hr 12-min integrations. RL rows report the final training episode and one frozen inference after ten training episodes. Dr Hook is disabled for the primary comparison and enabled only for the diagnostic RL row. The native control has no agent task, and the columns are overlapping task durations rather than additive components.}
\vspace{2mm}
\label{tbl:task-wall-times}
\begin{tabular}{lrrrr}
\toprule
Configuration & Training agent & Training UM & Inference agent & Inference UM\\
& (s) & (s) & (s) & (s)\\
\midrule
Native, non-nudged, Dr Hook off & --- & --- & --- & 112\\
Column-aware RL, Dr Hook off & 126 & 123 & 118 & 116\\
Diagnostic RL, Dr Hook on & 161 & 160 & 126 & 126\\
\bottomrule
\end{tabular}
\end{table}

Table~\ref{tbl:task-wall-times} reports the PBS task wall times for 6-hr 12-min integrations. The coupled inference UM task takes 116-s, 4-s (3.57\%) longer than the 112-s native task with the same executable and Dr Hook disabled. The UM and the RL agent start together, and their 116- and 118-s durations overlap and are not added. Across all ten training forecasts, individual UM task times range from 123 to 126-s. These are single-sequence observations, not repeatability estimates or end-to-end workflow latency. The implementation evaluates actions in batches of 128 columns, avoids redundant array copies, disables critic-parameter gradients during actor updates and defers unused replay-buffer imports. Learning updates use batches of 32 columns. For profiling, all 192 final actors and all six converted inference PP streams match a numerical reference using the same learning configuration byte for byte, preserving the reported forecast diagnostics.

\subsection{Inference runtime optimisation}
\label{app:runtime-optimisations}

Table~\ref{tbl:runtime-optimisations} traces the retained configurations using the complete inference UM task duration for a 6-hr 12-min forecast after ten training episodes. Both the inference and training histories use an old unoptimised column-aware experiment as a reference. The inference UM task took 197-s, while the overlapping agent task took 198-s. Runs affected by job-start delays are excluded from the comparison, and no estimated delay correction is applied.

Redis process parallelism accounts for the largest observed reduction: 194 to 140-s (54-s, 27.8\%). Each UM--agent rank pair is assigned explicitly to one of the independent Redis processes on the dedicated server node. 16 processes did not improve on eight, so the retained experiment uses 12. Removing an unused tensor exchange in the UM reduced incoming Redis traffic by approximately 5.34~GB per inference forecast.

The frozen-actor changes avoid constructing critics, target networks, optimisers and replay storage during inference, load only actor checkpoints and use the inference-only PyTorch execution context. Along with skipping training-object construction, this then reduced wall time to 127-s (13-s, 9.29\%), while mean rank-local initialisation fell from 11.774 to 0.0656-s. Disabling Dr Hook subsequently reduced 127 to 117-s through removal of diagnostic instrumentation rather than a change to forecast computation.

\begin{table}[!ht]
\centering
\small
\caption{Measured inference runtime along the retained optimisation sequence. Each reduction is relative to the preceding row: $100(t_{\mathrm{previous}}-t)/t_{\mathrm{previous}}$. Positive values mean a shorter runtime. Rows describe configuration changes, not isolated causal estimates, and bundled changes share one measured reduction. Dr Hook is enabled until the explicitly marked disabling step.}
\vspace{2mm}
\label{tbl:runtime-optimisations}
\begin{tabular}{p{0.61\linewidth}rrr}
\toprule
Configuration change & UM (s) & Cut (s) & Cut (\%)\\
\midrule
Column-aware original numerical reference & 197 & --- & ---\\
Remove unused tensor exchange & 194 & 3 & 1.52\\
Distribute ranks across 12 Redis processes & 138 & 56 & 28.87 \\
Streamline frozen-actor initialisation and execution & 127 & 11 & 7.97\\
Disable Dr Hook, retaining the same executable & 117 & 10 & 7.87\\
Use 128-column prediction batches, remove redundant copies and retain training-startup changes & 116 & 1 & 0.85\\
\midrule
Total reduction from the reference & 116 & 81 & 41.12\\
\bottomrule
\end{tabular}
\end{table}

Reducing prediction batches from 256 to 128 columns lowered mean cumulative inference prediction time from 3.123 to 2.127-s per rank (0.996-s, 31.89\%). Removing redundant tensor copies and action copy removal yielded a further 1-s reduction to 116-s. Training-specific changes disabled critic-parameter gradients during actor updates and deferred an unused replay-buffer import. In the second training episode, these respectively reduced the corresponding mean rank-local cumulative update time from 4.081 to 3.899-s (0.182-s, 4.46\%) and optimiser preload time from 13.223 to 3.145-s (10.078-s, 76.21\%). These component timings are not additive UM wall-time savings, and training-only changes are not credited with inference reductions.

The final ten-training-plus-one-inference confirmation took 116-s, a net 1-s reduction from the 117-s configuration and an overall 81-s reduction (41.12\%) from the 197-s reference. All 192 final actors and all six converted inference PP streams remained byte-identical to the ten-episode unoptimised numerical reference. The remaining overhead relative to the 112-s native control is 4-s (3.57\%). 

\subsection{Training runtime optimisation}
\label{app:training-optimisations}

Table~\ref{tbl:training-optimisations} traces individual final-training agent task durations from the original column-aware numerical reference to the final optimised configuration. Each measurement is the final nudged training forecast in a ten-training-plus-one-inference sequence, integrating 6-hr 12-min. The starting measurement is 208-s and the final measurement is 126-s for the agent, while the corresponding UM tasks take 203 and 123-s. These are per-forecast durations, not accumulated episode totals, and concurrent task durations are not added. 

\begin{table}[!ht]
\centering
\small
\caption{Individual final-training agent task durations across retained configurations, starting from the same column-aware reference as the inference history. The cut is relative to the preceding listed measurement: $100(t_{\mathrm{previous}}-t)/t_{\mathrm{previous}}$. Positive values mean shorter runtime, and negative values mean longer runtime. Dr Hook is enabled through the instrumented confirmation and disabled for the final configuration. }
\vspace{2mm}
\label{tbl:training-optimisations}
\begin{tabular}{p{0.60\linewidth}rrr}
\toprule
Configuration change & Agent (s) & Cut (s) & Cut (\%)\\
\midrule
Column-aware model with a single Redis process & 208 & --- & ---\\
Remove unused tensor exchange & 204 & 4 & 1.92\\
Use 12 independent Redis processes & 161 & 43 & 21.08 \\
Disable Dr Hook, curated replay, 128-column predictions, copy removal, critic-gradient gating and deferred replay import & 126 & 24 & 16.00 \\
\bottomrule
\end{tabular}
\end{table}

Redis process parallelism produced the largest early reduction: the individual training agent task fell from 204 to 161-s (43-s, 21.08\%). The final-training agent task took 126-s, 82-s (39.42\%) below the 208-s original column-aware reference. The final row is a combined configuration comparison and does not assign those 24 seconds to any one change. Direct curated-replay construction reduced mean rank-local construction time from 12.734 to 11.132-s (1.602-s, 12.58\%), critic-gradient disabling reduced second-forecast mean rank-local update time from 4.081 to 3.899-s (4.46\%), and deferring the unused replay-buffer import reduced optimiser preload from 13.223 to 3.145-s (76.21\%). All 192 final actors and the converted inference PP streams in the final ten-episode confirmation remained byte-identical to the ten-episode numerical reference.

\subsection{Instrumented coupling costs}

\begin{figure}[!h]
\centering
\includegraphics[width=\linewidth]{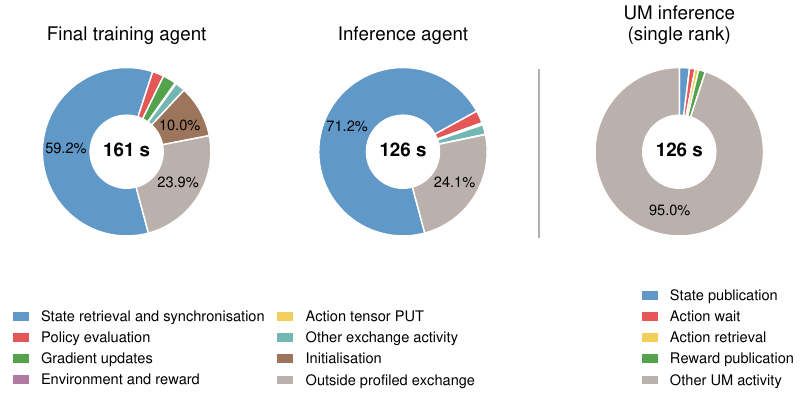}
\caption{Shares of complete task wall time in the separate Dr Hook-enabled diagnostic experiment for final nudged training, frozen non-nudged inference and the inference UM task. Agent durations are mean cumulative times across 192 ranks, without double-counting nested timers. The UM panel uses inclusive Dr Hook totals from MPI rank~96, with 31 state and reward publications and 93 action waits and retrievals across the three fields. Other UM activity is the residual of the 126-s task, not a direct timer. Only shares of at least 8\% are labelled.}
\label{fig:app-profiling-wall-time}
\end{figure}

In the diagnostic experiment, Figure~\ref{fig:app-profiling-wall-time} partitions complete task wall time into mutually exclusive components. Retrieval and synchronisation across the three fields account for 59.2\% of final-training agent time and 71.2\% of inference agent time. Policy evaluation contributes 2.2\% and 2.4\%, while learning contributes 2.6\% during training. Training initialisation, including the early optimiser-support import, contributes 10.0\%. The residual outside profiled exchange and initialisation is 23.9\% in training and 24.1\% in inference, including other startup, checkpoint and shutdown activity. On the selected UM rank, the four instrumented SmartRedis regions together account for 5.0\% of the 126-s inference task, with 95.0\% assigned to the residual outside those regions.

From Table~\ref{tbl:app-profiling}, the inclusive state-retrieval timer is much larger than the raw tensor GET because it includes waiting for the next UM state, rather than measuring transport alone. The full agent exchange also covers policy evaluation, tensor operations, reward construction and enabled learning. Its inference mean is 0.350-s shorter than in final training. The relatively small rank-to-rank spreads indicate broadly balanced average invocation costs, but they do not establish the absence of short-lived stragglers or variability between jobs.

The profiling tool Dr. Hook~\cite{saarinen_drhook_2005} separately measures the Fortran side on UM rank~96. Table~\ref{tbl:app-drhook-profiling} reports the mean time per invocation for the four instrumented SmartRedis regions during final training and inference. This selected-rank trace identifies where the synchronous UM process waits, but does not quantify variation among UM ranks. Its percentages cannot be interpreted as all-rank averages or as the entire difference from native runtime. These component timings describe the Dr Hook-enabled diagnostic configuration and do not provide a decomposition of the 116-s inference task. The matched Dr Hook-disabled comparison in Table~\ref{tbl:task-wall-times} measures additional UM elapsed time, while the separate agent and Redis allocations also contribute computational cost.

\clearpage

\begin{table}[!h]
\centering
\small
\caption{Rank-local Python wall-clock timings in the separate Dr Hook-enabled diagnostic experiment for the final nudged training forecast and frozen non-nudged inference. Values are the mean $\pm$ population standard deviation across 192 rank means, in seconds per invocation.\protect\footnotemark}
\vspace{2mm}
\label{tbl:app-profiling}
\begin{tabular}{
  l
  S[table-format=1.4(4),separate-uncertainty=true]
  S[table-format=1.4(4),separate-uncertainty=true]
}
\toprule
Component & {Final training} & {Inference}\\
& {Mean $\pm$ spread (s)} & {Mean $\pm$ spread (s)}\\
\midrule
Full agent exchange & 3.4336(674) & 3.0832(47)\\
Theta retrieval including synchronisation & 3.0243(677) & 2.8433(126)\\
U-wind retrieval including synchronisation & 0.0245(18) & 0.0249(12)\\
V-wind retrieval including synchronisation & 0.0249(21) & 0.0243(15)\\
Theta tensor GET & 0.0041(8) & 0.0042(6)\\
U-wind tensor GET & 0.0038(8) & 0.0039(6)\\
V-wind tensor GET & 0.0036(6) & 0.0035(6)\\
Policy evaluation & 0.1145(57) & 0.0991(47)\\
Theta action tensor PUT & 0.0014(2) & 0.0013(2)\\
U-wind action tensor PUT & 0.0013(2) & 0.0013(2)\\
V-wind action tensor PUT & 0.0012(2) & 0.0012(2)\\
Environment and reward step & 0.0070(8) & 0.0115(12)\\
Learning update (16 critic steps) & 0.1385(86) & {\text{---}}\\
Checkpoint save & 0.5776(2111) & {\text{---}}\\
\bottomrule
\end{tabular}
\end{table}
\footnotetext{The full-exchange and state-retrieval timers contain some of the other rows and therefore must not be added together.}

\begin{table}[!h]
\centering
\small
\caption{Dr Hook wall-clock timings in the diagnostic experiment on UM rank~96 for final nudged training and frozen non-nudged inference. Components are inclusive mean seconds per invocation. There are 31 state and reward publications and 93 action waits and retrievals in each forecast. The final row gives reported instrumentation overhead.}
\vspace{2mm}
\label{tbl:app-drhook-profiling}
\begin{tabular}{
  l
  S[table-format=1.4]
  S[table-format=1.4]
}
\toprule
Component & {Final training} & {Inference}\\
& {Mean time (s)} & {Mean time (s)}\\
\midrule
State publication & 0.0870 & 0.0793\\
Wait for agent action & 0.3274 & 0.0141\\
Action retrieval & 0.0106 & 0.0096\\
Reward-diagnostic publication & 0.0552 & 0.0538\\
Dr Hook overhead (\%) & 1.72 & 2.25\\
\bottomrule
\end{tabular}
\end{table}

\clearpage

\section{Additional spatial verification}
\label{app:spatial}
\setcounter{figure}{0}
\setcounter{table}{0}

\subsection{Reference diagnostics and expanded zonal MAE}
\label{sec:reference-diagnostics}

The verification reference is the UKMO analysis dump valid on 12 December 2021 at 06:00 UTC. Individual fields in that dump have different temporal definitions. $T_{500}$ and horizontal wind speed are derived from instantaneous model-level analysis fields at 06:00 UTC. $Z_{500}$ and MSLP use the stored 05:00--06:00 means, while $T_{1.5\textrm{m}}$ uses the stored maximum over the same hour. These diagnostics are compared with instantaneous +6-h forecast fields and both forecast configurations (nudged and non-nudged) use the same reference.

Temperature is obtained by interpolating rho-level Exner pressure to theta levels before conversion, followed by linear interpolation in pressure. For wind, analysis components and rho-level Exner pressure are regridded to the forecast wind grid, with matching model-level labels and heights checked before linear interpolation in pressure. Horizontal wind speed $W=\sqrt{u^2+v^2}$ is calculated after interpolating the components, and its MAE measures speed error rather than vector error. Analysis $Z_{250}$ is interpolated linearly in log pressure between the stored $Z_{200}$ and $Z_{300}$ fields, whereas $Z_{500}$ and $Z_{850}$ are selected directly. The forecast has all three levels explicitly. Near-surface temperature is evaluated at 1.5-m above local ground level.

Tables~\ref{tbl:extended-zonal-mae} and~\ref{tbl:extended-latitude-mae} report cosine-latitude-weighted pointwise MAE globally and in six disjoint latitude bands. Gain is the percentage reduction relative to native MAE: positive values mean improvement, negative values deterioration, and zero equal error. Each comparison uses cells finite in the native forecast, coupled forecast and analysis, excluding unbracketed pressure levels. These are spatial summaries of one initialisation, not independent forecast samples.

\begin{table}[!ht]
\centering\small
\caption{Global latitude-weighted MAE at +6~h for the non-nudged native and non-nudged coupled RL forecasts. Gain is $100(\mathrm{MAE}_{\mathrm{native}}-\mathrm{MAE}_{\mathrm{RL}})/\mathrm{MAE}_{\mathrm{native}}$. Positive means lower error (improvement), negative means higher error (deterioration), and zero means equal MAE.}
\vspace{2mm}
\label{tbl:extended-zonal-mae}
\begin{tabular}{llrrr}\toprule
 & & \multicolumn{3}{c}{Global}\\
\cmidrule(lr){3-5}
Variable & Units & Native MAE & RL MAE & Gain (\%)\\\midrule
$W_{250}$ & m~s$^{-1}$ & 0.84753 & 0.84766 & -0.02\\
$W_{500}$ & m~s$^{-1}$ & 0.73167 & 0.72973 & +0.27\\
$W_{850}$ & m~s$^{-1}$ & 0.74661 & 0.74573 & +0.12\\
$T_{250}$ & K & 0.23313 & 0.22273 & +4.46\\
$T_{500}$ & K & 0.23563 & 0.22347 & +5.16\\
$T_{850}$ & K & 0.41070 & 0.39129 & +4.73\\
$T_{1.5\mathrm{m}}$ & K & 0.53106 & 0.51901 & +2.27\\
$Z_{250}$ & m & 22.68236 & 23.24670 & -2.49\\
$Z_{500}$ & m & 2.70041 & 2.62342 & +2.85\\
$Z_{850}$ & m & 2.80846 & 2.71191 & +3.44\\
MSLP & hPa & 0.42518 & 0.41601 & +2.16\\
\bottomrule
\end{tabular}
\end{table}

\clearpage

\begin{table}[!ht]
\centering\footnotesize
\setlength{\tabcolsep}{3pt}
\caption{Zonal cosine-latitude-weighted MAE at +6~h on common finite cells. Each latitude band compares the non-nudged native control with non-nudged coupled RL using separate MAE and Gain (\%) subcolumns. Gain is $100(\mathrm{MAE}_{\mathrm{native}}-\mathrm{MAE}_{\mathrm{RL}})/\mathrm{MAE}_{\mathrm{native}}$: positive indicates improvement, negative deterioration, and zero equal error. Wind speed is in m~s$^{-1}$, temperature in K, height in m and MSLP in hPa. Pressure subscripts are hPa.}
\vspace{2mm}
\label{tbl:extended-latitude-mae}
\begin{tabular}{lrrrrrrrrr}\toprule
 & \multicolumn{3}{c}{60--90$^\circ$N} & \multicolumn{3}{c}{30--60$^\circ$N} & \multicolumn{3}{c}{0--30$^\circ$N}\\
\cmidrule(lr){2-4}\cmidrule(lr){5-7}\cmidrule(lr){8-10}
Variable & Native & RL & Gain (\%) & Native & RL & Gain (\%) & Native & RL & Gain (\%)\\\midrule
$W_{250}$ & 0.654 & 0.654 & -0.01 & 0.904 & 0.914 & -1.12 & 0.861 & 0.855 & +0.77\\
$W_{500}$ & 0.730 & 0.728 & +0.29 & 0.948 & 0.947 & +0.16 & 0.626 & 0.625 & +0.15\\
$W_{850}$ & 0.845 & 0.848 & -0.34 & 1.026 & 1.022 & +0.32 & 0.677 & 0.677 & +0.03\\
$T_{250}$ & 0.268 & 0.242 & +9.78 & 0.290 & 0.281 & +3.00 & 0.188 & 0.175 & +6.83\\
$T_{500}$ & 0.228 & 0.208 & +9.02 & 0.246 & 0.236 & +4.06 & 0.219 & 0.208 & +5.01\\
$T_{850}$ & 0.416 & 0.397 & +4.48 & 0.468 & 0.446 & +4.64 & 0.409 & 0.390 & +4.51\\
$T_{1.5\mathrm{m}}$ & 1.585 & 1.564 & +1.37 & 0.750 & 0.735 & +1.97 & 0.454 & 0.447 & +1.68\\
$Z_{250}$ & 6.908 & 6.676 & +3.35 & 17.071 & 17.060 & +0.06 & 29.619 & 30.205 & -1.98\\
$Z_{500}$ & 2.570 & 2.443 & +4.93 & 2.876 & 2.751 & +4.35 & 2.243 & 2.227 & +0.69\\
$Z_{850}$ & 2.888 & 2.550 & +11.68 & 2.929 & 2.863 & +2.25 & 2.498 & 2.473 & +0.98\\
MSLP & 0.471 & 0.449 & +4.52 & 0.543 & 0.539 & +0.69 & 0.333 & 0.341 & -2.63\\
\midrule & \multicolumn{3}{c}{0--30$^\circ$S} & \multicolumn{3}{c}{30--60$^\circ$S} & \multicolumn{3}{c}{60--90$^\circ$S}\\
\cmidrule(lr){2-4}\cmidrule(lr){5-7}\cmidrule(lr){8-10}
Variable & Native & RL & Gain (\%) & Native & RL & Gain (\%) & Native & RL & Gain (\%)\\\midrule
$W_{250}$ & 0.903 & 0.903 & -0.07 & 0.830 & 0.832 & -0.23 & 0.677 & 0.669 & +1.26\\
$W_{500}$ & 0.661 & 0.657 & +0.69 & 0.769 & 0.772 & -0.48 & 0.697 & 0.684 & +1.73\\
$W_{850}$ & 0.642 & 0.637 & +0.77 & 0.694 & 0.697 & -0.45 & 0.725 & 0.730 & -0.69\\
$T_{250}$ & 0.187 & 0.178 & +4.77 & 0.243 & 0.240 & +1.18 & 0.357 & 0.341 & +4.54\\
$T_{500}$ & 0.241 & 0.231 & +4.38 & 0.233 & 0.226 & +2.96 & 0.261 & 0.227 & +13.15\\
$T_{850}$ & 0.436 & 0.408 & +6.33 & 0.351 & 0.344 & +2.22 & 0.312 & 0.292 & +6.34\\
$T_{1.5\mathrm{m}}$ & 0.381 & 0.370 & +2.87 & 0.267 & 0.251 & +5.92 & 0.448 & 0.442 & +1.24\\
$Z_{250}$ & 30.438 & 31.537 & -3.61 & 19.051 & 19.800 & -3.93 & 8.874 & 9.229 & -4.00\\
$Z_{500}$ & 2.403 & 2.248 & +6.44 & 2.806 & 2.804 & +0.07 & 4.879 & 4.838 & +0.83\\
$Z_{850}$ & 3.173 & 2.951 & +6.99 & 2.348 & 2.308 & +1.70 & 3.459 & 3.564 & -3.03\\
MSLP & 0.412 & 0.385 & +6.53 & 0.341 & 0.327 & +4.06 & 0.681 & 0.682 & -0.05\\
\bottomrule
\end{tabular}
\end{table}

Global temperature MAE decreases by 4.46\%, 5.16\% and 4.73\% at 250, 500 and 850~hPa, respectively, with improvement in every latitude band at each level. Wind-speed gains are smaller and regionally mixed: $-0.02\%$, +0.27\% and +0.12\% at those levels. Geopotential height MAE decreases by 2.85\% at 500~hPa and 3.44\% at 850~hPa, but increases by 2.49\% at 250~hPa. These results extend the variable coverage of the feasibility evaluation without establishing generalisation to other dates.

\clearpage

\subsection{Spatial error and bias diagnostics}

\begin{figure}[!h]
\centering
\includegraphics[width=0.94\linewidth]{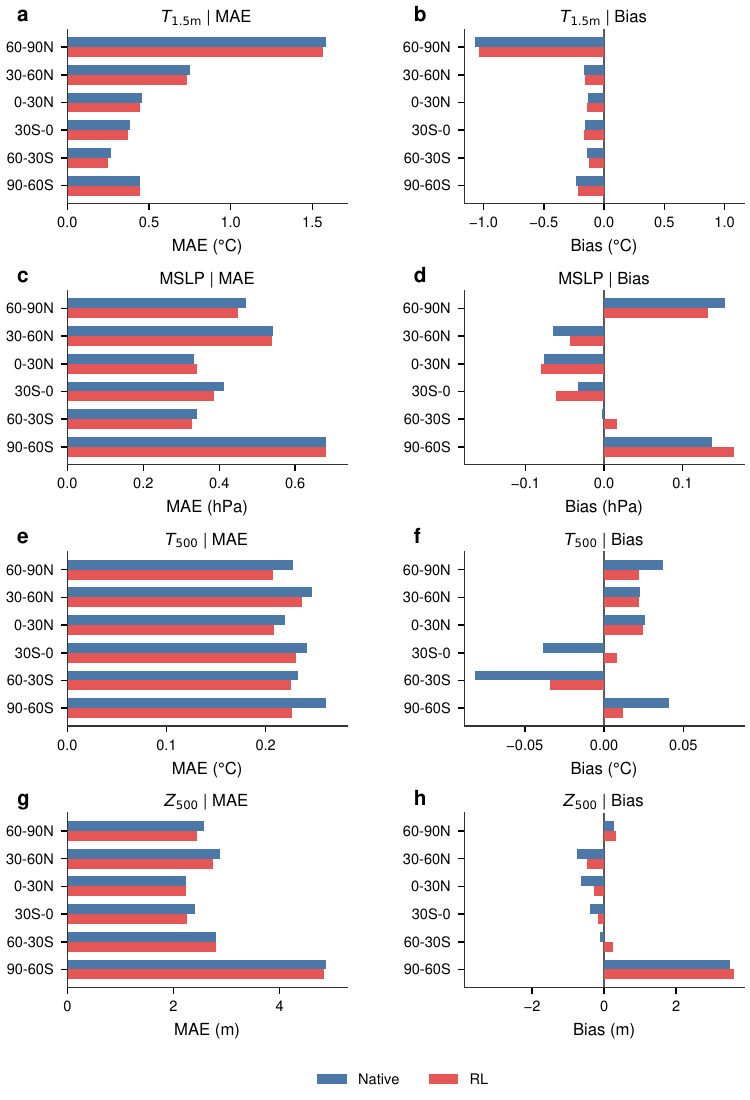}
\caption{Latitude-weighted zonal verification at +6~h against UKMO analysis. Panels a, c, e and g compare non-nudged native and coupled RL MAE for $T_{1.5\mathrm{m}}$, MSLP, $T_{500}$ and $Z_{500}$. Panels b, d, f and h show signed biases. Each bar is a cosine-latitude-weighted band mean. Temperature and $Z_{500}$ MAE decrease in all six bands, while MSLP improves in four. Biases do not consistently move towards zero. Analysis time aggregation follows Section~\ref{sec:reference-diagnostics}. These summaries describe one initialisation, not independent forecasts.}
\label{fig:app-zonal-metrics}
\end{figure}

\clearpage

\begin{figure}[!h]
\centering
\makebox[\linewidth][c]{\includegraphics[width=1.15\linewidth]{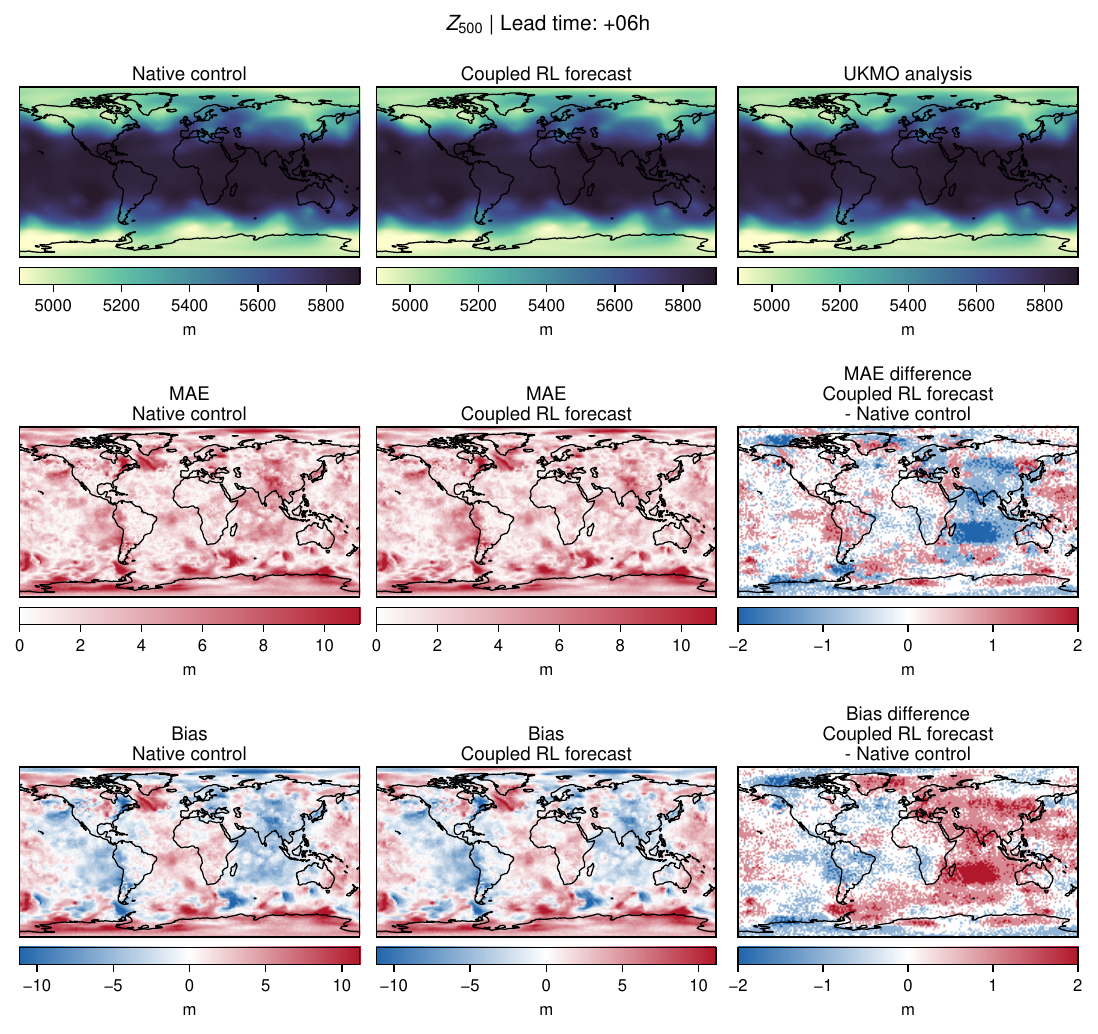}}
\caption{$Z_{500}$ verification at +6~h against the stored 05:00--06:00 analysis mean. The top row compares non-nudged native, coupled RL and UKMO analysis. The middle row shows pointwise absolute errors and their RL-minus-native difference. The bottom row shows signed errors and their difference. Blue in the middle-right panel indicates reduced absolute error, whereas blue in the bottom-right panel indicates a more negative signed error, not necessarily a smaller error. Colour limits use the 1st--99th percentiles for fields and the 99th percentile of error magnitudes.}
\label{fig:app-z500-spatial}
\end{figure}

\clearpage

\begin{figure}[!h]
\centering
\makebox[\linewidth][c]{\includegraphics[width=1.15\linewidth]{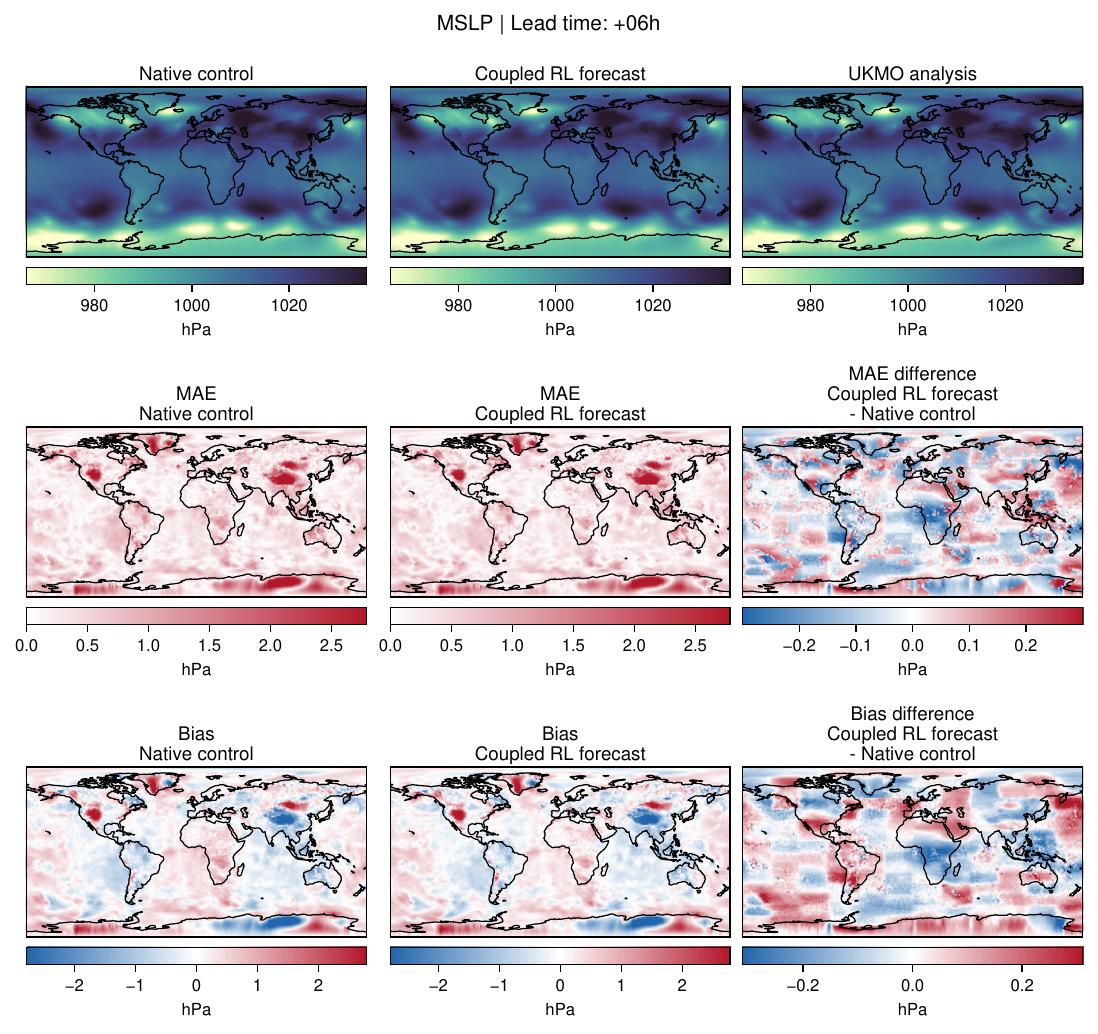}}
\caption{MSLP verification at +6~h against the stored 05:00--06:00 analysis mean. Rows show the non-nudged native forecast, coupled RL forecast and analysis, then pointwise absolute errors and signed errors, with RL-minus-native differences in the rightmost column. Colours and percentile limits follow Figure~\ref{fig:app-z500-spatial}, with all values in hPa.}
\label{fig:app-mslp-spatial}
\end{figure}

\clearpage
\begin{figure}[!h]
\centering
\makebox[\linewidth][c]{\includegraphics[width=1.15\linewidth]{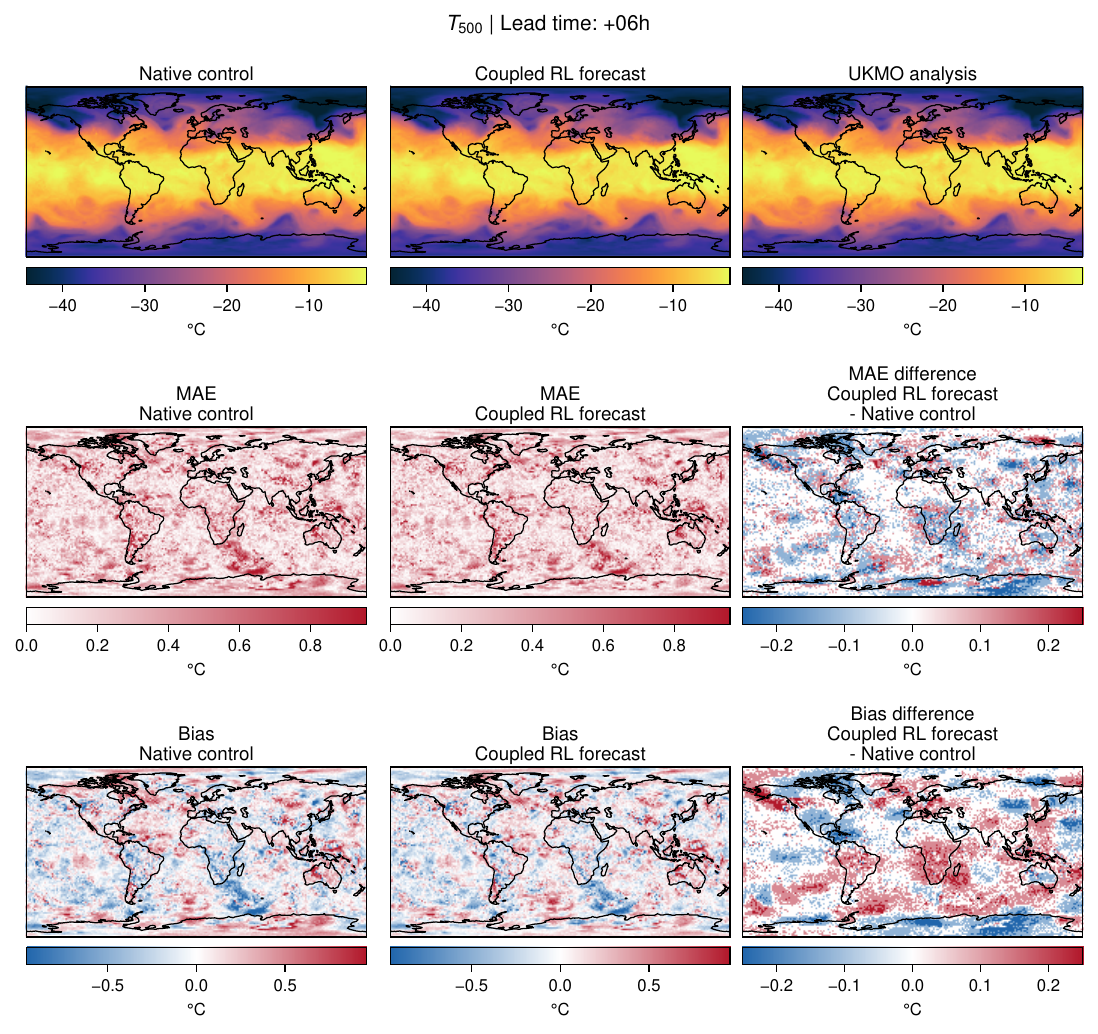}}
\caption{$T_{500}$ verification at +6~h, using the same non-nudged controls, rows and colour conventions as Figure~\ref{fig:app-z500-spatial}. Analysis temperature is derived by colocating Exner pressure with potential temperature before conversion and interpolation to 500~hPa. The absolute-error and signed-error differences are not equivalent.}
\label{fig:app-t500-spatial}
\end{figure}

\clearpage

\begin{figure}[!h]
\centering
\makebox[\linewidth][c]{\includegraphics[width=1.15\linewidth]{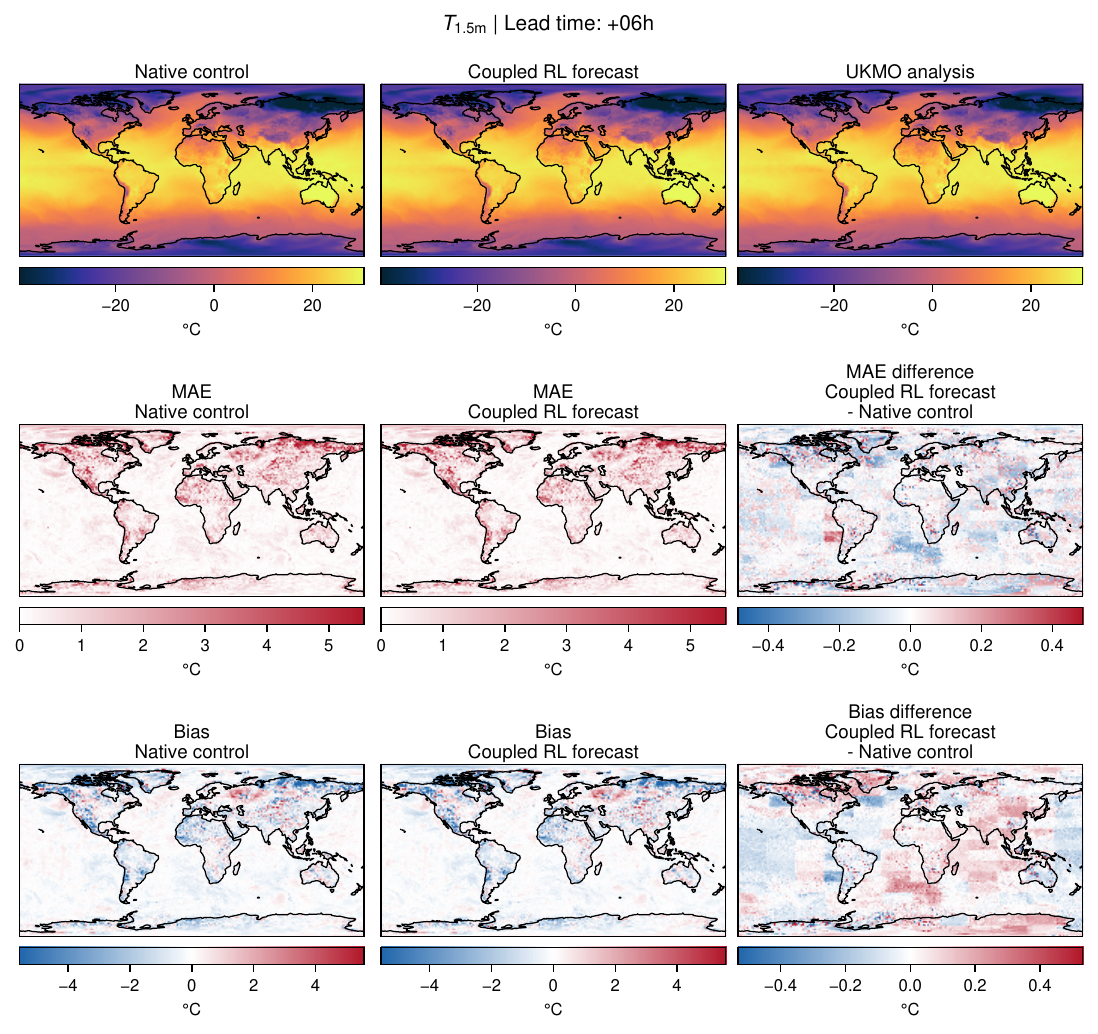}}
\caption{$T_{1.5\mathrm{m}}$ verification at +6~h against the stored 05:00--06:00 analysis maximum. The top row shows non-nudged native, coupled RL and analysis fields. The middle row shows absolute errors and their RL-minus-native difference, and the bottom row shows signed errors and their difference. Colour conventions and percentile limits follow Figure~\ref{fig:app-z500-spatial}.}
\label{fig:app-t1p5m-spatial}
\end{figure}

\clearpage
\section{Comparison with the analysis-nudged control}
\label{app:nudged-reference}
\setcounter{table}{0}

The analysis-nudged native forecast provides a separate reference for learning the effects of nudging. Unlike the frozen RL forecast, this control applies analysis information throughout integration, so it is not an information-matched operational forecast comparison. Both are verified at +6~h against the supplied UKMO diagnostics, retaining the temporal definitions and 250-hPa height interpolation described in Section~\ref{sec:reference-diagnostics}. Tables~\ref{tbl:nudged-global} and~\ref{tbl:nudged-zonal} give global and zonal MAEs for all eleven diagnostics. Here gain is the percentage reduction relative to the nudged native MAE: positive values favour the non-nudged RL forecast, and negative values favour the nudged control.

\begin{table}[!ht]
\centering\small
\caption{Global latitude-weighted MAE at +6~h for the analysis-nudged native and non-nudged coupled RL forecasts. Gain is $100(\mathrm{MAE}_{\mathrm{native}}-\mathrm{MAE}_{\mathrm{RL}})/\mathrm{MAE}_{\mathrm{native}}$: positive means lower error (improvement), negative means higher error (deterioration), and zero means equal MAE. }
\vspace{2mm}
\label{tbl:nudged-global}
\begin{tabular}{llrrr}\toprule
 & & \multicolumn{3}{c}{Global}\\
\cmidrule(lr){3-5}
Variable & Units & Nudged MAE & RL MAE & Gain (\%)\\\midrule
$W_{250}$ & m~s$^{-1}$ & 0.76373 & 0.84766 & -10.99\\
$W_{500}$ & m~s$^{-1}$ & 0.64620 & 0.72973 & -12.93\\
$W_{850}$ & m~s$^{-1}$ & 0.62998 & 0.74573 & -18.37\\
$T_{250}$ & K & 0.19807 & 0.22273 & -12.45\\
$T_{500}$ & K & 0.20058 & 0.22347 & -11.41\\
$T_{850}$ & K & 0.32384 & 0.39129 & -20.83\\
$T_{1.5\mathrm{m}}$ & K & 0.49306 & 0.51901 & -5.26\\
$Z_{250}$ & m & 22.56368 & 23.24670 & -3.03\\
$Z_{500}$ & m & 3.51100 & 2.62342 & +25.28\\
$Z_{850}$ & m & 3.28857 & 2.71191 & +17.54\\
MSLP & hPa & 0.46651 & 0.41601 & +10.82\\
\bottomrule
\end{tabular}
\end{table}

\clearpage

\begin{table}[!ht]
\centering\footnotesize
\setlength{\tabcolsep}{3pt}
\caption{Zonal cosine-latitude-weighted MAE at +6~h on common finite cells. Each latitude band compares the analysis-nudged native control with non-nudged coupled RL using separate MAE and Gain (\%) subcolumns. Gain is $100(\mathrm{MAE}_{\mathrm{native}}-\mathrm{MAE}_{\mathrm{RL}})/\mathrm{MAE}_{\mathrm{native}}$: positive indicates improvement, negative deterioration, and zero equal error. Wind speed is in m~s$^{-1}$, temperature in K, height in m and MSLP in hPa. Pressure subscripts are hPa. }
\vspace{2mm}
\label{tbl:nudged-zonal}
\begin{tabular}{lrrrrrrrrr}\toprule
 & \multicolumn{3}{c}{60--90$^\circ$N} & \multicolumn{3}{c}{30--60$^\circ$N} & \multicolumn{3}{c}{0--30$^\circ$N}\\
\cmidrule(lr){2-4}\cmidrule(lr){5-7}\cmidrule(lr){8-10}
Variable & Nudged & RL & Gain (\%) & Nudged & RL & Gain (\%) & Nudged & RL & Gain (\%)\\\midrule
$W_{250}$ & 0.567 & 0.654 & -15.26 & 0.939 & 0.914 & +2.67 & 0.680 & 0.855 & -25.73\\
$W_{500}$ & 0.608 & 0.728 & -19.76 & 0.939 & 0.947 & -0.76 & 0.485 & 0.625 & -28.84\\
$W_{850}$ & 0.704 & 0.848 & -20.39 & 0.915 & 1.022 & -11.78 & 0.530 & 0.677 & -27.74\\
$T_{250}$ & 0.212 & 0.242 & -13.78 & 0.268 & 0.281 & -4.94 & 0.145 & 0.175 & -20.74\\
$T_{500}$ & 0.177 & 0.208 & -17.24 & 0.241 & 0.236 & +2.04 & 0.171 & 0.208 & -21.87\\
$T_{850}$ & 0.304 & 0.397 & -30.78 & 0.376 & 0.446 & -18.78 & 0.314 & 0.390 & -24.24\\
$T_{1.5\mathrm{m}}$ & 1.566 & 1.564 & +0.16 & 0.693 & 0.735 & -6.06 & 0.422 & 0.447 & -5.90\\
$Z_{250}$ & 6.245 & 6.676 & -6.91 & 17.042 & 17.060 & -0.10 & 29.169 & 30.205 & -3.55\\
$Z_{500}$ & 2.072 & 2.443 & -17.92 & 3.065 & 2.751 & +10.25 & 3.978 & 2.227 & +44.00\\
$Z_{850}$ & 2.312 & 2.550 & -10.31 & 3.068 & 2.863 & +6.69 & 3.891 & 2.473 & +36.43\\
MSLP & 0.400 & 0.449 & -12.19 & 0.558 & 0.539 & +3.47 & 0.473 & 0.341 & +27.85\\
\midrule & \multicolumn{3}{c}{0--30$^\circ$S} & \multicolumn{3}{c}{30--60$^\circ$S} & \multicolumn{3}{c}{60--90$^\circ$S}\\
\cmidrule(lr){2-4}\cmidrule(lr){5-7}\cmidrule(lr){8-10}
Variable & Nudged & RL & Gain (\%) & Nudged & RL & Gain (\%) & Nudged & RL & Gain (\%)\\\midrule
$W_{250}$ & 0.698 & 0.903 & -29.46 & 0.896 & 0.832 & +7.15 & 0.680 & 0.669 & +1.69\\
$W_{500}$ & 0.513 & 0.657 & -28.14 & 0.770 & 0.772 & -0.33 & 0.645 & 0.684 & -6.05\\
$W_{850}$ & 0.502 & 0.637 & -27.05 & 0.640 & 0.697 & -8.96 & 0.665 & 0.730 & -9.70\\
$T_{250}$ & 0.139 & 0.178 & -27.93 & 0.242 & 0.240 & +0.82 & 0.290 & 0.341 & -17.51\\
$T_{500}$ & 0.189 & 0.231 & -21.86 & 0.221 & 0.226 & -2.17 & 0.209 & 0.227 & -8.39\\
$T_{850}$ & 0.334 & 0.408 & -22.31 & 0.301 & 0.344 & -14.19 & 0.256 & 0.292 & -14.07\\
$T_{1.5\mathrm{m}}$ & 0.338 & 0.370 & -9.44 & 0.232 & 0.251 & -8.18 & 0.431 & 0.442 & -2.56\\
$Z_{250}$ & 30.572 & 31.537 & -3.16 & 19.415 & 19.800 & -1.98 & 8.029 & 9.229 & -14.94\\
$Z_{500}$ & 3.871 & 2.248 & +41.91 & 3.146 & 2.804 & +10.86 & 4.081 & 4.838 & -18.55\\
$Z_{850}$ & 3.730 & 2.951 & +20.90 & 2.590 & 2.308 & +10.90 & 2.880 & 3.564 & -23.76\\
MSLP & 0.449 & 0.385 & +14.20 & 0.346 & 0.327 & +5.49 & 0.650 & 0.682 & -4.93\\
\bottomrule
\end{tabular}
\end{table}

The RL forecast has smaller global $Z_{500}$, $Z_{850}$ and MSLP errors than this control, but larger temperature and wind-speed errors at all three pressure levels, larger $T_{1.5\textrm{m}}$ error, and larger error against the interpolated $Z_{250}$ reference. Both polar bands also retain larger $Z_{500}$ and MSLP errors. Improvement over non-nudged native execution in the main text should not be interpreted as uniformly reproducing or exceeding analysis nudging.

\end{document}